\documentclass[lettersize,journal,twoside]{IEEEtran}
\IEEEoverridecommandlockouts 
\usepackage{graphicx}
\usepackage{threeparttable}
\usepackage{multirow} 
\usepackage{hyperref}
\usepackage{url}
\usepackage{caption}
\usepackage{algorithm}
\usepackage{algorithmic}
\usepackage{amsmath}
\usepackage{booktabs}
\usepackage{amssymb}
\usepackage{xcolor}
\usepackage{comment}
\usepackage[capitalize]{cleveref}
\crefname{section}{Sec.}{Secs.}
\Crefname{section}{Section}{Sections}
\Crefname{table}{Table}{Tables}
\crefname{table}{Tab.}{Tabs.}
\makeatletter
\usepackage{xspace}
\DeclareRobustCommand\onedot{\futurelet\@let@token\@onedot}
\def\@onedot{\ifx\@let@token.\else.\null\fi\xspace}

\def\etal{{et al}\onedot}
\makeatother

\title{\LARGE \bf
UGNA-VPR: A Novel Training Paradigm for Visual Place Recognition Based on Uncertainty-Guided NeRF Augmentation
}

\author{Yehui Shen, Lei Zhang, Qingqiu Li,  Xiongwei Zhao, Yue Wang, Huimin Lu, Xieyuanli Chen  
\thanks{Manuscript received: January 5, 2025; Revised: February 24, 2025; Accepted: March 16, 2025. This paper was recommended for publication by Editor \mbox{Abhinav Valada} upon evaluation of the Associate Editor and Reviewers’ comments.}
\thanks{Y. Shen, H. Lu and X. Chen are with College of Intelligence Science and Technology, National University of Defense Technology, China. Y. Shen is additional with Faculty of Robot Science and Engineering, Northeastern University, China. X. Zhao is with School of Electronic and Information Engineering, Harbin Institute of Technology (Shenzhen), China. Y. Wang is with State Key Laboratory of Industrial Control Technology and Institute of Cyber-Systems and Control, Zhejiang University, China. L. Zhang is with SIASUN Robot\& Automation Co., Ltd, China. Q. Li is with School of Computer Science, Fudan University, China.}%
\thanks{This work was supported in part by the National Science Foundation of China (Grant No. 62403478), Young Elite Scientists Sponsorship Program by CAST (No. 2023QNRC001), and  Major Project of Natural Science Foundation of Hunan Province (Grant No. 2021JC0004).}
\thanks{Digital Object Identifier (DOI): see top of this page.}
}

\begin{document}

\maketitle


\begin{abstract}
Visual place recognition (VPR) is crucial for robots to identify previously visited locations, playing an important role in autonomous navigation in both indoor and outdoor environments. However, most existing VPR datasets are limited to single-viewpoint scenarios, leading to reduced recognition accuracy, particularly in multi-directional driving or feature-sparse scenes. Moreover, obtaining additional data to mitigate these limitations is often expensive.
This paper introduces a novel training paradigm to improve the performance of existing VPR networks by enhancing multi-view diversity within current datasets through uncertainty estimation and NeRF-based data augmentation. Specifically, we initially train NeRF using the existing VPR dataset. Then, our devised self-supervised uncertainty estimation network identifies places with high uncertainty. The poses of these uncertain places are input into NeRF to generate new synthetic observations for further training of VPR networks. Additionally, we propose an improved storage method for efficient organization of augmented and original training data. 
We conducted extensive experiments on three datasets and tested three different VPR backbone networks. The results demonstrate that our proposed training paradigm significantly improves VPR performance by fully utilizing existing data, outperforming other training approaches. We further validated the effectiveness of our approach on self-recorded indoor and outdoor datasets, consistently demonstrating superior results. Our dataset and code have been released at  \href{https://github.com/nubot-nudt/UGNA-VPR}{https://github.com/nubot-nudt/UGNA-VPR}.
\end{abstract}

\section{Introduction}
Visual place recognition (VPR) plays a critical role in applications such as autonomous robot navigation and self-driving vehicles~\cite{yin2024survey}. The VPR process typically first compresses a query image into a descriptor, which is then matched against descriptors from previously visited locations stored in a database. By identifying descriptor similarities, the system determines if the robot has returned to a previously visited place, enabling essential downstream tasks such as loop closing and relocalization. Current VPR methods remain highly challenging due to the difficulty of generating robust descriptors from limited-viewpoint observational data. 

To address this problem, some works have developed various neural networks to generate robust descriptors~\cite{ali2023mixvpr, wang2022transvpr}, while others have focused on collecting more data to enhance VPR~\cite{pepperell2014icra, warburg2020mapillary}. Although both approaches can enhance VPR performance, designing an ideal network is challenging, while acquiring and maintaining additional data is costly. Few studies have explored the efficient use of existing data for VPR improvement. Some data augmentation techniques like flipping, translation, and rotation have been applied~\cite{yu2017giscience}. However, these methods are not specifically designed for VPR and may not fully exploit the potential of existing VPR datasets.
Recently, neural radiance field (NeRF)~\cite{mildenhall2021nerf} have demonstrated remarkable performance in synthetic view rendering, and has been applied to enhance various visual tasks, such as pose regress~\cite{chen2022dfnet}, localization~\cite{kuang2023ral} and SLAM~\cite{rosinol2023iros}. These applications suggest potential benefits of NeRF for VPR but have not been fully explored yet, especially in terms of how to evaluate the impact of each augmented data on VPR, enabling selective data augmentation to improve VPR performance.

\begin{figure}[t]
  \centering
  \includegraphics[width=1\columnwidth]{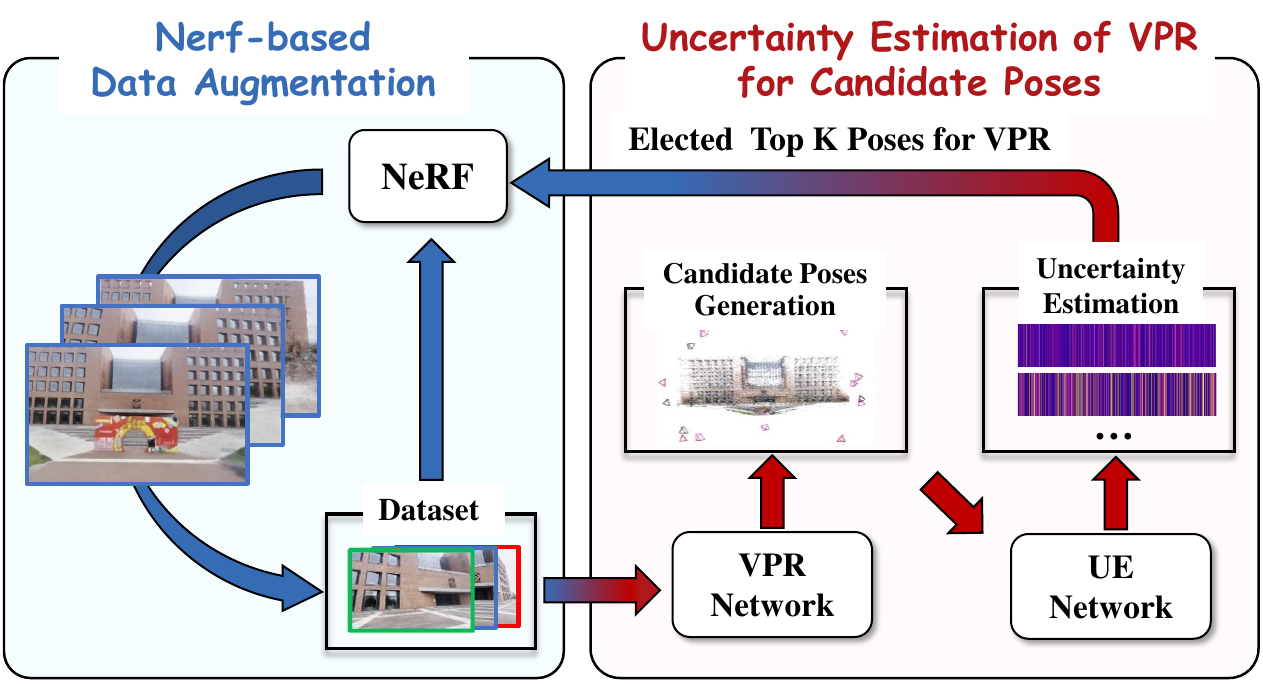}
  \captionsetup{aboveskip=2pt, belowskip=0pt}
  \caption{Overview of our novel VPR training paradigm, which consists of two main components: NeRF-based data augmentation and uncertainty estimation for selecting candidate rendering poses. We first use existing datasets to train both the NeRF and VPR networks. Based on the performance of the VPR network, a set of candidate rendering poses is generated. The uncertainty estimation network then assesses the uncertainty in descriptors of these candidate poses. The candidates with high uncertainties are selected and rendered by NeRF. The generated synthetic images are added to the VPR network training to enhance VPR performance.}
  \vspace{-0.2cm}
  \label{fig:workflow}
\end{figure}

In this paper, we propose a novel VPR training paradigm named UGNA-VPR that uses NeRF to generate new viewpoint observations based on our VPR uncertainty estimation, thereby enhancing VPR training and performance. Our key insight is that NeRF can effectively learn 3D models from limited 2D images, enabling the generation of synthetic novel views for poorly VPR-learned places, thus augmenting training and improving VPR performance.
Specifically, we propose a VPR uncertainty estimation network to evaluate the quality of place learning by a VPR network. Our method then uses NeRF to selectively generate synthetic views at different poses with high uncertainties near this place and incorporates these views into the dataset to enhance VPR model training. To efficiently organize both real and synthetic data, We also propose an effective data organization method for VPR training. The overview of our method is shown in~\cref{fig:workflow}. Extensive experiments have been conducted to evaluate the effectiveness of our proposed training paradigm and data organization method on both public and self-recorded datasets using three different VPR networks. These results demonstrate that our method significantly enhances VPR performance compared to other state-of-the-art baseline training methods and is compatible with various existing VPR techniques. 

In summary, the contributions of this paper are fourfold:
\begin{itemize}
    \item We propose UGNA-VPR, a novel training paradigm for VPR based on uncertainty-guided NeRF augmentation, which enhances the effectiveness of existing VPR models without the need for additional data collection.
    \item We design a self-supervised uncertainty estimation network, which can assess VPR uncertainty for a given place pose. This allows the effective selection of poses for NeRF rendering to augment VPR training data.
    \item We propose an improved data organization method for training, which efficiently organizes both real and synthetic data, enhancing the effectiveness of our NeRF-based augmentation method.
    \item We collected indoor and outdoor datasets, suitable for both 3D reconstruction and VPR. We have released our dataset and code to support the research community.
\end{itemize}

\section{Related Work}
This paper focuses on enhancing VPR performance using uncertainty-guided NeRF augmentation. Thus, we discuss three categories of related works: place recognition methods, NeRF-based data augmentation, and uncertainty estimation.

\subsubsection{Place Recognition} Much of the early work for VPR relied on hand-crafted descriptors to identify the most suitable candidate places in a database~\cite{vysotska2019effective}. With the advent of neural networks, deep learning-based methods rapidly advanced. R. Arandjelovic~\etal~\cite{arandjelovic2016netvlad} proposed NetVLAD, which used convolutional neural networks to extract features through an end-to-end training approach, achieving significant success in VPR. Subsequently, various NetVLAD-based methods emerged~\cite{hausler2021patch, li2023hot}. The introduction of Transformers~\cite{vaswani2017attention} marked a new peak in the development of deep learning-based VPR methods, with numerous researchers enhancing VPR performance by improving network architectures~\cite{ma2023cvtnet, peng2024transloc4d}. Recently, the emergence of large model technology has highlighted the importance of data~\cite{oquab2023dinov2}, leading researchers to enhance VPR effectiveness by leveraging massive datasets~\cite{keetha2023anyloc}.

\subsubsection{NeRF-based Data Augmentation} Neural radiance fields (NeRF) is a neural network-based 3D modeling technique~\cite{mildenhall2021nerf}. Due to its ability to generate high-quality 3D scene and object renderings, some works utilized it for data augmentation to improve the effectiveness of various tasks~\cite{li2023pac, feldmann2024nerfmentation}. Zhou~\etal~\cite{zhou2023nerf} employed NeRF for data augmentation to refine robotics strategies using hand-eye cameras. Chen~\etal~\cite{chen2022dfnet} proposed DFNet, which leverages NeRF for data augmentation to enhance pose regression. These methods often employ random data generation for augmentation, resulting in excessive data and wasted storage space. Our approach tackles this by introducing an uncertainty estimation module to selectively choose data, thereby improving data utilization efficiency while achieving better enhancements.

\subsubsection{Uncertainty Estimation} Kendall and Gal~\cite{kendall2017uncertainties} provide a theoretical foundation for evaluating model uncertainty in semantic segmentation, demonstrating that such assessments can enhance model effectiveness. NeU-NBV~\cite{jin2023neunbv} uses uncertainty estimation to guide view planning in unknown scenarios. In VPR, some works~\cite{cai2022iros,shen2024icra} leverage knowledge distillation to enhance the performance of the student model by evaluating the uncertainty in the teacher model. Unlike them, we do not modify the original VPR network. Instead, we employ uncertainty estimation to evaluate the uncertainty of newly added training data related to the current VPR network. This method enhances data augmentation, thereby enhancing the effectiveness of VPR.


\begin{figure*}[t]
  \centering
\includegraphics[width=0.85\linewidth]{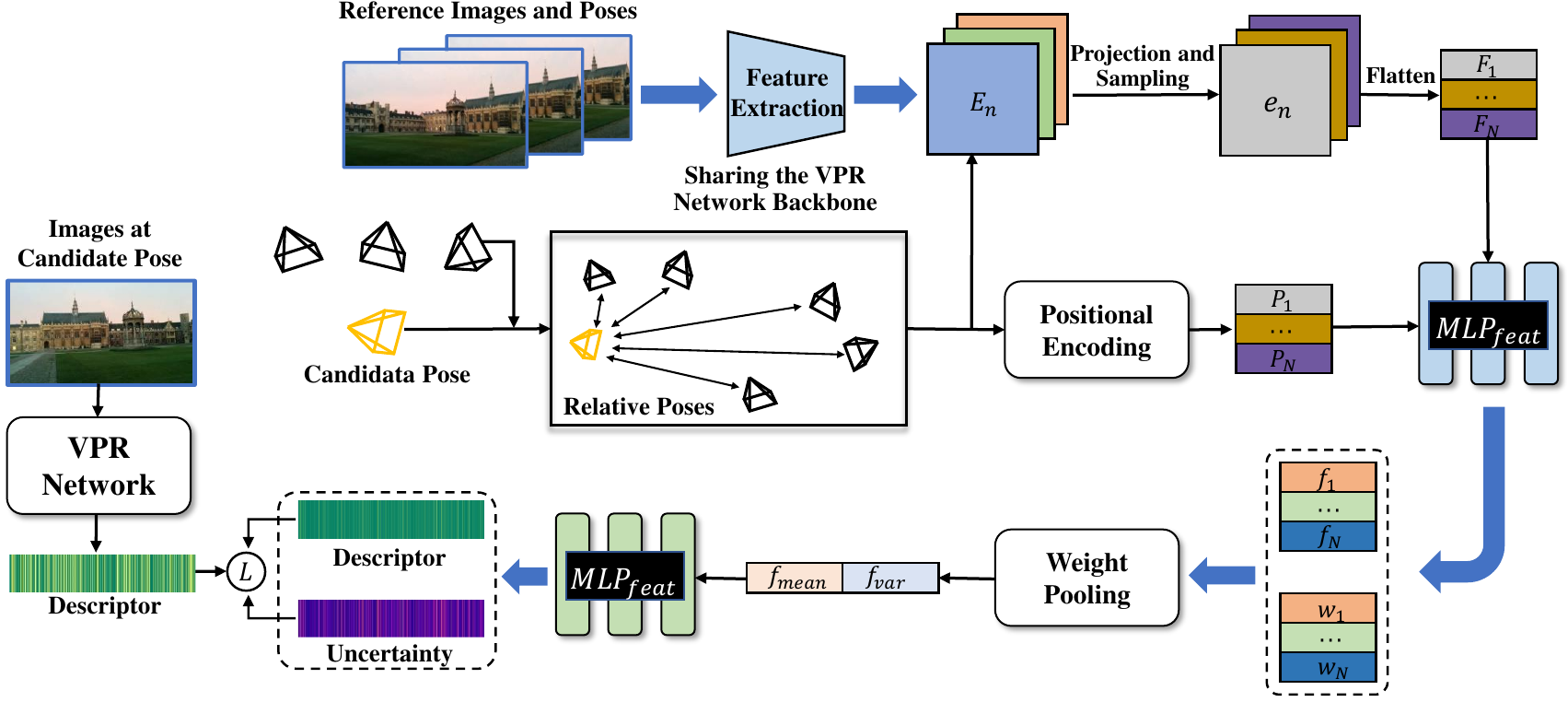}
\captionsetup{aboveskip=2pt, belowskip=0pt}
  \caption{The training pipeline for our uncertainty estimation (UE) network. The network is trained using real data and estimates the uncertainty of VPR descriptors generated by each candidate pose, which is used for the synthetic data selection.}
  \label{fig:unNet}
\vspace{-0.6cm}
\end{figure*}

\section{Methodology}
{We propose a training paradigm based on uncertainty-guided NeRF augmentation. The procedure of our method is illustrated in~\cref{fig:workflow}. First, we train the VPR and NeRF networks using the existing real dataset. The VPR network is then used to identify places in that dataset where the model fails to correctly recognize. Then, we generate several candidate rendering poses near these poorly performing places. For each candidate pose, our uncertainty estimation (UE) network estimates the VPR uncertainty based on co-visibility consistency from nearby reference images. Candidates with high uncertainty are then used for NeRF rendering, and the generated synthetic images are added to the dataset to further train the VPR network. 
The following subsections present the details of each module in our method.

\subsection{Uncertainty Estimation Network}
Our method does not modify the architecture of existing VPR networks but introduces an uncertainty estimation (UE) network to identify places that require additional synthetic observations rendered from NeRF to improve VPR performance. 
The structure of our proposed UE network primarily consists of a feature extraction backbone shared with the VPR network, and two MLP networks for position fusion as illustrated in~\cref{fig:unNet}. 
We first train the VPR network using the limited existing dataset and identify VPR failure cases within it. 
For each failure case, we randomly sample $M$ candidate rendering poses near the corresponding failure pose. Only the top $K$ uncertain candidate poses are selected to generate synthetic observations for further VPR training.
Unlike typical approaches that estimate VPR uncertainty directly from real images~\cite{cai2022iros}, we estimate the uncertainty using the candidate image pose with nearby reference images and poses before NeRF rendering. By doing so, it avoids rendering massive candidate images, that could be resource-consuming. 

To achieve this, we choose $N$ nearest images and their corresponding poses in the vicinity of each VPR failure place as reference images $I_{n\in\{1,2,..N\}}$ and poses.  
Our UE network uses the same backbone $V$ shared with the VPR network to extract the features of each reference image 
$E_n = V(I_n)\in\mathbb{R}^{H{\times}W{\times}C}$, where
$H$, $W$, and $C$ represent the feature height, width, and channels. The relative pose $\mathbf{T}_{ic}\!=\!(\mathbf{p_{ic}},\mathbf{r_{ic}})$ between the i-th reference image and the candidate are computed for later transforming the image features from reference to the candidate rendering pose, where $\mathbf{p_{ic}}$ denotes the translation and $\mathbf{r_{ic}}$ is the rotation. To obtain higher frequency details in the features, we encode $\mathbf{p_{ic}}$ into a higher-dimensional space using the position encoding operation proposed by Mildenhall~\etal~\cite{mildenhall2021nerf}, and combine it with the rotation $\mathbf{r_{ic}}$ to construct the pose features $P_{n\in\{1,2,..N\}}$, where $n$ denotes the combination of the information of the $n$-th reference image pose.

To allow the candidate to capture feature information from the reference images, we perform a projection on the reference image features based on their relative poses. 
\begin{equation}
\vspace{-2pt}
O_i=\mathbf{T}_{ic}(x_i,y_i,0,1)^\top ,
\vspace{-2pt}
\end{equation}
where $(x_i,y_i)$ denotes the coordinates of a point on the feature plane of the i-th reference image, while $O_i$ is the homogeneous coordinate of the projection of that point onto the plane of the candidate pose.

We then use bilinear interpolation to sample and obtain the new corresponding reference image features
$e_n\in\mathbb{R}^{H{\times}W{\times}C}$ from $E_n$. Next, the image features are flattened as $F_{n\in\{1,2,..N\}}$, which is aligned with the previously obtained pose features $P$. Then, the image feature $F_n$ and the pose feature $P_n$ are fed together into the $\text{MLP}_{\text{feat}}$~\cite{yu2021pixelnerf}, which fuse appreance information from image feature $F_n$ and spatial information from pose feature $P_n$ to obtain the new feature $f_{n\in\{1,2,..d\}}$, and outputs the predictive weights $W_{n\in\{1,2,..d\}}$. We use the predictive weights $W_n$ to compute the feature mean $f_{\text{mean}}$ and variation $f_{\text{var}}$, aggregating the information of all reference image features: 
\begin{equation}
\vspace{-2pt}
f_{\text{mean}k}= \frac{1}{N} \sum^{\substack{N}}_{\substack{i=1}}\left(f_{ik}{\times}W_{ik}\right) ,
\vspace{-2pt}
\end{equation}
\begin{equation}
\vspace{-2pt}
f_{\text{var}}= \frac{1}{N-1} \sum^{\substack{N}}_{\substack{i=1}}\left(f_{ik}{\times}W_{ik}-f_{\text{mean}k}\right)^2 ,
\vspace{-2pt}
\end{equation}
where, $i\in{N}$ represents the i-th reference image, $k\in{d}$ represents the k-th dimensional feature of the reference image feature $f_{n\in\{1,2,..d\}}$.
Finally, we use the $\text{MLP}_{\text{out}}$ to interpret a fused descriptor of the candidate pose $f_{\text{mean}}$ and its uncertainty descriptor $f_{\text{var}}$.
Note that, our UE module is a general framework, thus compatible with various VPR backbones. In this paper, we provide examples of using MixVPR~\cite{ali2023mixvpr}, CricaVPR~\cite{lu2024cricavpr}, and EigenPlaces~\cite{Berton2023EigenPlaces}. 

\begin{figure}[t]
  \centering
  \includegraphics[width=0.95\linewidth]{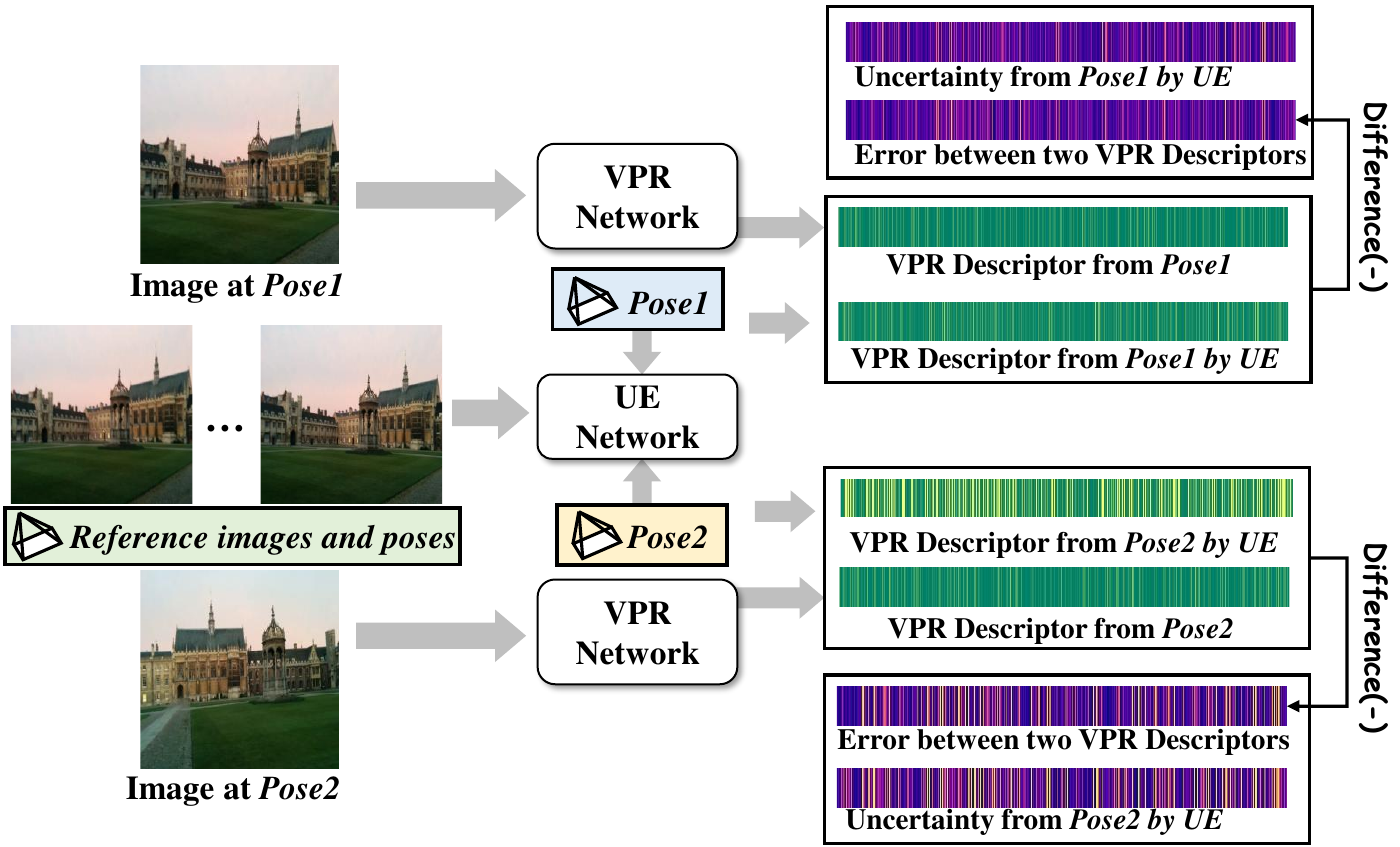}
  \captionsetup{aboveskip=2pt, belowskip=0pt}
  \caption{Examples of uncertainty estimation. With the same reference images, our UE network estimates the uncertainty for two candidate poses. Brighter regions in the uncertainty descriptors indicate that the uncertainty in the descriptors for pose 2 is higher than that for pose 1. Consequently, the difference between the VPR descriptors generated by the uncertainty pose and those generated by the corresponding image from the VPR network is greater for pose 2.}
  \label{fig:des}
 \vspace{-0.4cm}
\end{figure}

\subsection{Self-supervised Training for UE Network}
To guide the UE network in learning meaningful candidate pose descriptors and uncertainty, we propose a self-supervised training scheme that utilizes all collected data.
We want each dimension of our uncertainty descriptor to reflect the uncertainty associated with each dimension of the VPR descriptor, as each dimension corresponds to different image pixels with varying levels of uncertainty.
Therefore, we model each dimension $d_i\{i=1,2...D\}$ of the predicted VPR descriptors as an independent logistic normal distribution by normalizing the values of the predicted VPR descriptors between $\left(0,1\right)$ as:
\begin{equation} 
\begin{split}
p(d_i; \mu_i, \sigma_i) = 
\frac{1}{\sigma_i \sqrt{2\pi}} \frac{1}{d_i\left(1 - d_i\right)} 
e^{\left( -\frac{\left(\text{logit}\left(d_i\right) - \mu_i\right)^2}{2\sigma_i^2}\right)} ,
\end{split}
\end{equation}
where, $\text{logit}\left(d_i\right)=ln\left(\frac{d_i}{1-d_i}\right){\sim} N(\mu_i,\sigma_i^2)$ satisfies the standard normal distribution, with the mean $\mu_i$ and variance $\sigma_i$ predicted by our network. 

As introduced by Kendall and Gal~\cite{kendall2017uncertainties}, the negative log-likelihood can be used to learn aleatoric uncertainty by evaluating the alignment between model predictions and true values in regression tasks. During training, we minimize the negative log-likelihood $-\text{log}p\left(d_i=y_i|\mu_i,\sigma_i\right)$ between descriptors generated by our UE network and those generated by the original VPR network. 
Given the value $y_i$ of each dimension of the VPR descriptor and the value $d_i$ of the descriptor obtained by our UE network, we can calculate the uncertainty loss for each dimension as:
\begin{equation} 
\begin{split}
L=\sum^{\substack{N}}_{\substack{i=1}}\frac{1}{2}\text{log}\left(\sigma_i^2\right)+\text{log}\left(y_i\left(1 - y_i\right)\right)
+\frac{\left(\text{logit}\left(y_i\right)-\mu_i\right)^2}{2\sigma_i^2} .
\end{split}
\end{equation}

To steer the model's output away from extreme values, we introduce $\text{log}\left(y_i\left(1 - y_i\right)\right)$
as a penalty term for predictions of extreme probabilities. This loss measures the similarity between UE estimated descriptors and original VPR descriptors. Since UE estimated descriptors are the fused and transformed reference descriptors, it basically measures the co-visibility consistency between the features of nearby reference images and the query image to model the uncertainty of the VPR network. 

During deployment, given a candidate rendering pose along with reference images, our network can then predict the mean $u_i$ and variance $\sigma_i$, assuming that the VPR descriptor dimensions follow a normal distribution in the logit space. 
The mean and variance values are treated as the descriptor and uncertainty values for each dimension of our UE descriptor and uncertainty.

\begin{figure}[t]
  \centering
  \includegraphics[width=0.95\columnwidth]{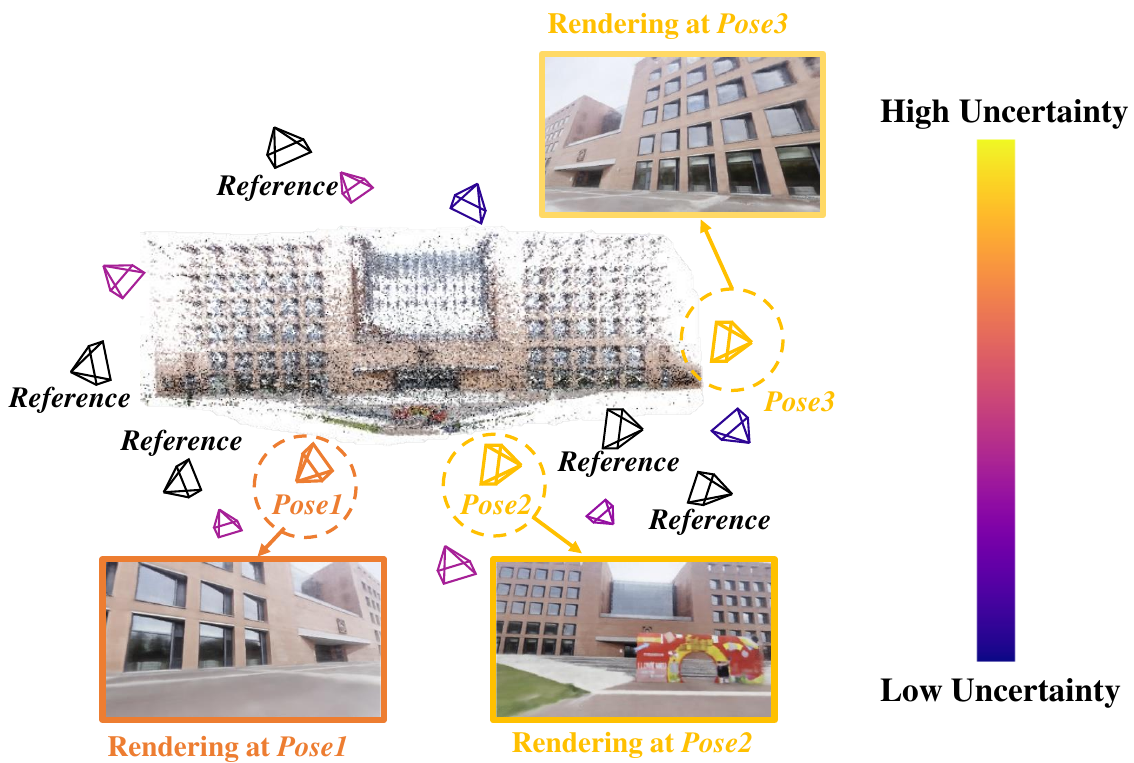}
  \captionsetup{aboveskip=2pt, belowskip=0pt}
  \caption{Uncertainty estimation during VPR training. Using the same reference information, we employ an uncertainty estimation network to evaluate the uncertainty in descriptors of candidate poses and select those with high uncertainty to render for subsequent training.}
  \label{fig:view1}
  \vspace{-0.6cm}
\end{figure}

An example is illustrated in~\cref{fig:des}, given the same reference images and poses, the uncertainty descriptor of a candidate Pose2 generated by the UE network is higher compared to a candidate Pose1. Correspondingly, the difference between the VPR descriptor of Pose2 inferred by the UE network and that generated by the VPR network with the corresponding image is also larger. 
This demonstrates that our UE network can learn to estimate the uncertainty of candidate poses in a self-supervised manner, without the need for manual labels.

\subsection{VPR Training Paradigm and Data Organization}
Once our self-supervised UE model is trained, we can use it to infer the uncertainty of candidate poses before rendering images, determining whether the data from that pose is informative enough to be added to VPR training.
As shown in~\cref{fig:view1}, after completing an epoch of VPR training, we evaluate the trained VPR model using existing query images in the dataset. To enable the VPR network to learn from challenging instances and improve robustness in poorly performing areas without overfitting to regions that have been well-learned, we adopt the following strategy: If the current VPR network fails to correctly identify the location for a query image (ImageQ) with the corresponding pose (poseQ), we randomly generate $M$ new candidate poses near the poseQ. Enhancing data around failure poses helps the VPR network learn from challenging instances, improving robustness in poorly performing areas without overfitting to already well-learned regions.Using the $N$ nearest images and their corresponding poses of ImageQ as reference images and poses, we estimate the uncertainty of these newly generated candidate poses via our UE module. We then select the poses with the top $K$ uncertainty, generate corresponding images using NeRF, and add these generated images along with their associated poses to the dataset to continue training the VPR network. This process helps improve the VPR network's performance by leveraging high-uncertainty poses to enrich the training data. Note that our UE model and NeRF are used only during VPR training and do not impact real-time VPR performance in practical deployment.
Theoretically, any synthesis rendering method could be used in our approach, and we use two different NeRF networks~\cite{chen2022dfnet,martin2021nerf} in experiments as examples. 

During the training phase, VPR datasets are typically divided into two parts: the database set and the query set. In traditional training methods, both sets consist solely of real images. In this paper, our approach augments the original real images with corresponding multi-view synthetic images. In real-world applications, the query set must consist of real images, as these represent the robot's real-time observed images during operation. For the database set, if synthetic images are directly mixed with some real images, the number of triplet training samples remains determined by the number of query images. This limits the size of the VPR training dataset and constrains improvements in VPR network accuracy.
To address this, we propose using all real images as query images while constructing the database only with synthetic images. This strategy maximizes the training samples, enables the generation of more triplets for VPR training, and optimizes the use of the limited dataset. Our experiments show a significant improvement in VPR performance with this real and synthetic data organization method.

Notably, using all real images as queries is only feasible because our uncertainty-guided NeRF augmentation method can generate novel observations of the same place from different viewpoints. This makes our data organization approach and uncertainty-guided NeRF augmentation method an effective solution for enhancing VPR performance with limited datasets.




\section{Experiments and Analyses}
We present our experiments to show the capabilities of our method and support our key claims that: 
(i) Our UGNA-VPR training paradigm, based on uncertainty-guided NeRF augmentation, improves the performance of existing VPR networks without the need to modify the original VPR networks and collect additional data.
(ii) Our uncertainty estimation network selects valuable candidate rendering poses for VPR training, introducing beneficial synthetic observations that enhance the VPR network training.
(iii) Our data organization method effectively leverages limited real images and generated synthetic data to enhance the VPR networks training.

\subsection{Experimental Setup}
\subsubsection{Dataset} 
We evaluated UGNA-VPR on the Cambridge (Cam)~\cite{kendall2015posenet}, Library (LIB), and Conference Room (CON) datasets. The Cam dataset is a publicly available outdoor dataset containing five scenes. The LIB and CON datasets were collected by us. The LIB dataset is an outdoor dataset containing five scenes from a campus environment, recorded at different times and dates, with a training/validation/test split of 2448/1223/3669 frames. The CON dataset is an indoor dataset consisting of five room scenes, also recorded at different times and dates, with a training/validation/test split of 2551/1272/3821 frames. The poses for both datasets were obtained using a NeRF with SLAM software, nerfstudio~\cite{nerfstudio}. 

\subsubsection{Configuration}
We selected three state-of-the-art VPR networks as VPR backbones to demonstrate the suitability of our proposed training paradigm across various existing VPR approaches: MixVPR~\cite{ali2023mixvpr}, CricaVPR~\cite{lu2024cricavpr}, and EigenPlaces~\cite{Berton2023EigenPlaces}. We pre-trained them on the given datasets with the image resolution of $224\times224$, and using the default VPR descriptors with 512 dimensions. To prevent overfitting, we employed early stopping criteria during training. For VPR networks, training was halted if no improvement was observed on the validation set for 10 consecutive epochs. 
We also tested our NeRF augmentation method with two different NeRF methods, NeRF-O~\cite{mildenhall2021nerf} and NeRF-H~\cite{chen2022dfnet}. 

In the training phase of the VPR network, we adhered to existing settings~\cite{shen2024icra,cai2022iros} with 8 batches, using the Adam optimizer with an initial learning rate of $1\!\times\!10^{-5}$, decaying 0.99 times after each calendar element, and a weight decay of 0.001. In the training phase of the uncertainty estimation network, we adhere to the NBV setting~\cite{jin2023neunbv}, using the Adam optimizer with an initial learning rate of $1\!\times\!10^{-5}$, decaying by a factor of 0.999 after each calendar element. To restrict memory consumption, we selected the nearest five images as reference images for training. Our Uncertainty estimation training infrastructure is hosted on a cloud server equipped with a single RTX 4090(24GB) GPU. Our VPR training infrastructure is hosted on a server equipped with a single L20(48GB) GPU.

\subsubsection{Evaluation Metrics} Consistent with other approaches, we use Recall@N as an evaluation metric to assess the effectiveness of our method and other training paradigms. Recall@N measures a query image successfully retrieved if at least one of the top N candidates is correctly identified. We present Recall@1 in all tables within the main text, while additional Recall@N results are shown in Fig.~\ref{fig:VPR_performance}. 

\subsubsection{Baseline Training Paradigms}
We compare our methods with five baseline training paradigms,
The Regular-Training refers to the standard training paradigm that does not employ any data augmentation techniques. 
Yu~\etal~\cite{yu2017giscience} applied conventional data augmentation techniques, such as rotation and flipping, to increase the data before training. Zhou~\etal~\cite{zhou2023nerf} augments data before training by randomly rendering images using NeRF. 
Chen~\etal~\cite{chen2022dfnet} uses NeRF to randomly render images during each epoch of training, but the images rendered in one epoch are not retained for the next. 
Feldmann~\etal~\cite{feldmann2024nerfmentation} also uses NeRF to randomly generate rendered images during each epoch of training, but the images rendered in each epoch are retained until the end of the training process. For fair comparisons, the amounts of additional data generated by different approaches are the same, as well as our method. 

\begin{table*}[t]
\centering
\renewcommand\arraystretch{1.2}
\renewcommand\tabcolsep{7pt}
\captionsetup{aboveskip=0pt, belowskip=0pt}
\caption{VPR performance with different backbone and datasets}
\begin{tabular}{l|ccc|ccc|ccc}\hline 
     &\multicolumn{3}{|c|}{MixVPR~\cite{ali2023mixvpr}}    &\multicolumn{3}{|c|}{EigenPlaces~\cite{Berton2023EigenPlaces}}   &\multicolumn{3}{|c}{CricaVPR~\cite{lu2024cricavpr}}                       \\ \hline
Training Paradigm & Cambridge    & LIB  & CON  & Cambridge     & LIB  & CON   & Cambridge    & LIB  & CON    \\
    \hline
Regular-Training  &67.50  &89.19 &91.67  &81.24  &92.30  &94.73  &53.08  &88.04  &89.54  \\
Yu~\etal~\cite{yu2017giscience}  &68.50  &89.35  &91.75 &81.45  &93.37  &94.81 &53.24  &87.14  &90.34\\
Zhou~\etal~\cite{zhou2023nerf}  &68.39 &89.52  &92.14 &81.24 &92.79  &93.87 &53.03 &86.73  &90.33 \\
Chen~\etal~\cite{chen2022dfnet} &68.29 &89.36  &92.06   &81.87  &92.55  &94.50  &52.35  &87.14 &90.17   \\
Feldmann~\etal~\cite{feldmann2024nerfmentation}  &68.34 &88.86  &91.82 &82.18 &92.30 &94.18 &53.81  &86.49  &90.33 \\
Ours-regular &\underline{68.70} &\underline{90.17} &\underline{92.53}  &\underline{82.60} &\underline{93.45} &\underline{95.20} &\underline{54.49} &\underline{88.12} &\textbf{90.64} \\
Ours &\textbf{72.05} &\textbf{93.20}  &\textbf{93.71}    &\textbf{89.81} &\textbf{93.69} &\textbf{95.28}  &\textbf{61.60} &\textbf{88.53} &\textbf{90.64} \\   \hline
\end{tabular}
\begin{tablenotes} 
\centering
\item \textbf{bold} denotes the best and \underline{underline} denotes the second best
\end{tablenotes}
  \label{tab:MixVPR}
   \vspace{-0.4cm}
\end{table*}

\begin{figure*}[t]
  \centering
  \includegraphics[width=0.95\linewidth]{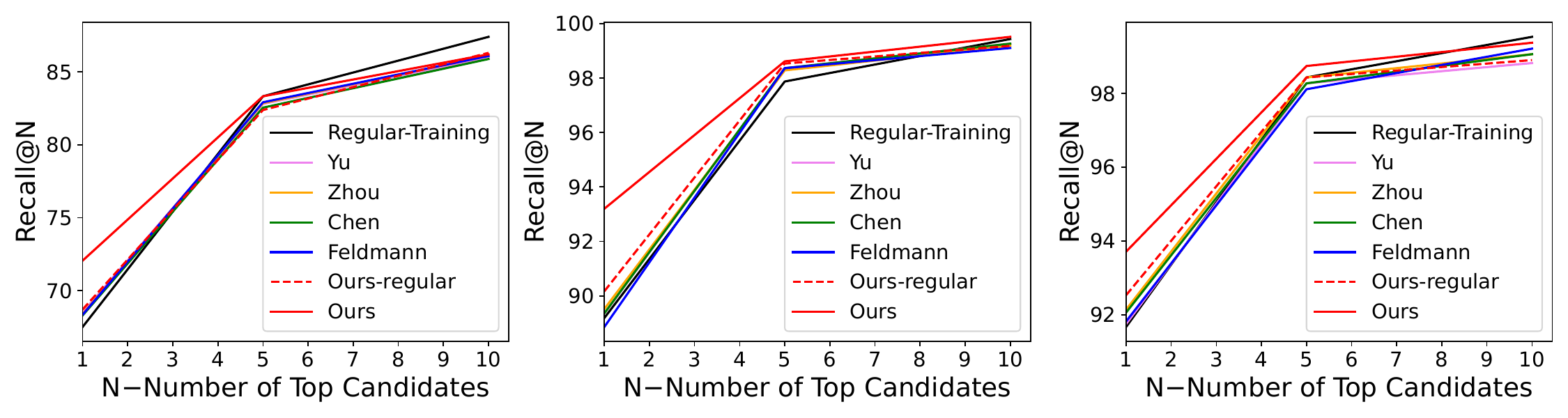}
  \captionsetup{aboveskip=2pt, belowskip=0pt}
  \caption{Recall@N on different datasets with MixVPR. (a) Cambridge. (b) LIB. (c) CON.}
  \label{fig:recall_MixVPR}
  \vspace{-0.4cm}
\end{figure*}

\begin{figure*}[t]
\centering
\includegraphics[width=0.95\linewidth]{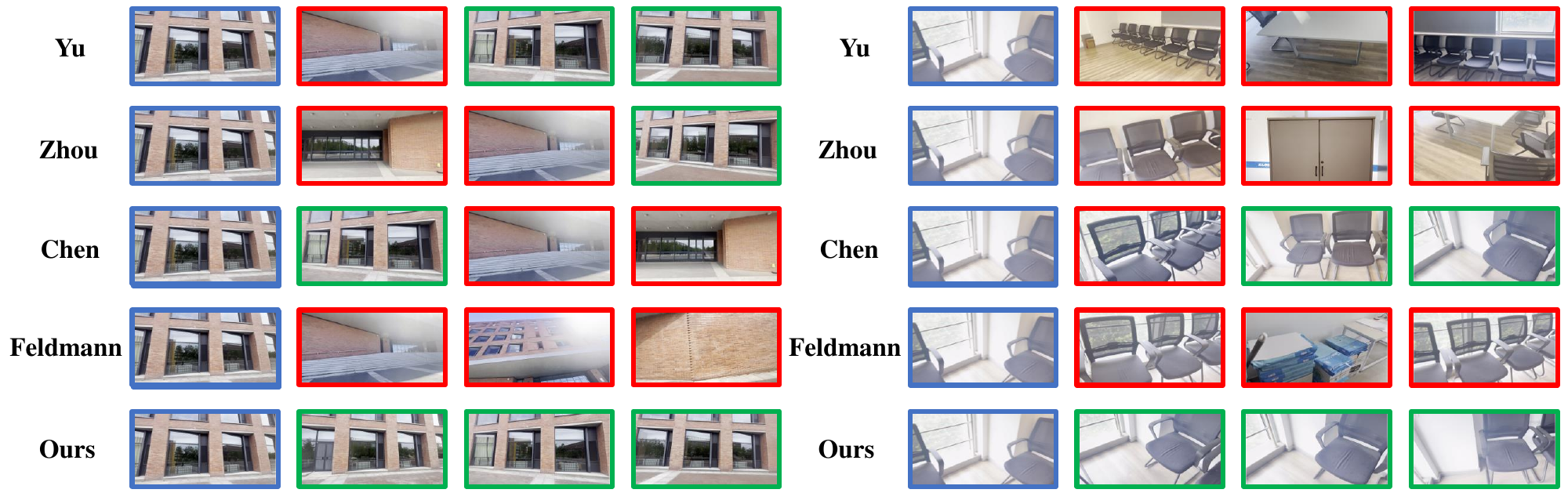}
\captionsetup{aboveskip=2pt, belowskip=0pt}
  \caption{Qualitative VPR results comparison using different baseline training paradigms. The blue represents the query, the green represents the correct query results, and the red represents the incorrect query results.}
  \label{fig:VPR_performance}
  \vspace{-0.5cm}
\end{figure*}
\vspace{-0.1cm}




\subsection{VPR Performance Improvements}
The first experiment aims to demonstrate the effectiveness of our proposed training paradigm and data organization method. Due to computational resource constraints, we add three selected rendering poses for each VPR failure place during each epoch. For instance, if there are $L$ poses in the dataset queries that fail to correctly identify the top one pose in a given epoch, we generate $L\!\times\!3$ images and poses for that epoch and add them to the dataset database. We show two versions of ours; the first one, Ours-regular, uses UGNA-VPR with the same data organization method as that used by the other baseline methods, and the second one, Ours, uses UGNA-VPR plus the proposed data organization method. 

\cref{tab:MixVPR} presents the VPR Recall@1 performance on various datasets using different VPR networks, including MixVPR, EigenPlaces and CricaVPR. Cambridge and LIB are outdoor datasets, and CON is an indoor dataset. It can be observed that our method consistently outperforms other baseline training paradigms in enhancing VPR performance of different VPR networks across both outdoor and indoor environments. This demonstrates its strong generalization ability and effectiveness in improving various VPR networks.
Even with standard data organization, UGNA-VPR surpasses other training paradigms. When combined with our proposed real and synthetic data organization approach, VPR performance is further enhanced, demonstrating the effectiveness and good match of our data organization approach and UGNA-VPR.
Notably, in challenging scenarios like CricaVPR of Cambridge dataset, our approach improves the Recall@1 metric by more than 8\%, highlighting the impact of our augmentation training paradigm. As shown, MixVPR achieves better performance across the three datasets compared to the other two backbones, as its smaller parameter size allows it to benefit the most from our method.

\cref{fig:recall_MixVPR} shows the Recall@N results for all methods on different datasets. As shown, our method consistently enhances VPR performance across different datasets, outperforming other training paradigms in terms of Recall@N. The fewer the retrieved candidates, the harder the task, where our method significantly improves VPR performance under such cases. When retrieving ten candidates, the recalls of all methods are similar. \cref{fig:VPR_performance} presents the qualitative VPR results of our method compared to different baseline training paradigms, demonstrating superior improvements in retrieval capabilities.


\begin{table*}[t]
\centering
\renewcommand\arraystretch{1.2}
\renewcommand\tabcolsep{7pt}
\captionsetup{aboveskip=0pt, belowskip=0pt}
\caption{Effect of uncertainty estimation with different backbone on different datasets.}
\begin{tabular}{cc|ccc|ccc|ccc}
\hline
 & &\multicolumn{3}{|c|}{MixVPR~\cite{ali2023mixvpr}}    &\multicolumn{3}{|c|}{EigenPlaces~\cite{Berton2023EigenPlaces}}   &\multicolumn{3}{|c}{CricaVPR~\cite{lu2024cricavpr}}                       \\ \hline
Data generated pre-epochs &UE  &Cambridge &LIB &CON &Cambridge &LIB &CON &Cambridge &LIB &CON \\
\hline
 &         &68.50  &89.52  &92.14 &82.08 &93.12  &95.13 &51.57  &87.71  &90.09 \\

    &$\surd$   &\textbf{68.70} &\textbf{90.17} &\textbf{92.53}  &\textbf{82.60} &\textbf{93.45} &\textbf{95.20} &\textbf{54.49} &\textbf{88.12} &\textbf{90.64}  \\
\hline 
$\surd$ &  &67.92  &89.27  &91.98 &81.89  &93.20  &94.34  &53.66 &87.31  &89.54 \\

$\surd$ &$\surd$  &\textbf{68.44} &\textbf{89.84}  &\textbf{92.06} &\textbf{82.50} &\textbf{93.28} &\textbf{94.65} &\textbf{55.59} &\textbf{87.71} &\textbf{90.02} \\
\hline
\end{tabular}
  \label{tab:UE on MixVPR}
    \vspace{-0.3cm}
\end{table*}

\subsection{Ablation Studies and Insights}
\subsubsection{Uncertainty Estimation}
This experiment evaluates the impact of our proposed UE module. Besides, we also analyze whether to maintain the previous epochs (pre-epochs) generated synthetic data. As shown in \cref{tab:UE on MixVPR}, we conducted experiments with four different setups: with and without keeping the data generated by the previous epochs, and with and without using our UE network for candidate poses. 
The results demonstrate that our proposed UE module consistently delivers good results across various environments and different VPR network backbones, regardless of whether data generated by previous epochs is retained. These results highlight the superior performance and robust generalization capability of our UE network.
Maintaining the synthetic data generated from the previous epoch does not significantly enhance performance, likely because the network has already effectively learned the information within previous epochs.

\subsubsection{Number of Candidate Rendering Poses}
This study analyzes the impact of the number $M$ of chosen candidate rendering poses on improving VPR performance. As shown in~\cref{tab: MixVPR diferent M}, adding rendering poses (ours) generally yields better results than not adding any synthetic observations (regular). However, increasing the number of candidate rendering poses does not necessarily lead to improved performance. This may be because, as 
$M$ increases, the likelihood of selecting candidate poses that do not contribute novel information from the NeRF model also increases. These rendered images may contain little to no useful information or even introduce noise, and incorporating a large number of such images into the dataset could degrade VPR performance.

\subsubsection{Study on Data Organization}
Besides validating our data organization method (Our-Data-Org) with our training paradigm, we also applied it to the baseline methods 
in~\cref{tab:data}. Here, we use the MixVPR as the backbone and experiment on three datasets. The results show that our data organization method also enhances other NeRF-based training paradigms, while our UGNA-VPR plus our data organization method (Ours) still yields the best performance.


\begin{table}[t]
\centering
\renewcommand\arraystretch{1.2}
\renewcommand\tabcolsep{8pt}
\captionsetup{aboveskip=-1pt, belowskip=-1pt}
\caption{Recognition performance with MixVPR with different $M$. }
\begin{tabular}{c|ccc}
\hline
$M$   &Cambridge &LIB &CON  \\ \hline
0   &67.50  &89.19 &91.67     \\
10  &68.34 &89.76 &92.45  \\
20  &\textbf{68.70} &\textbf{90.17} &\textbf{92.53}   \\
30 &68.23 &89.76 &92.45   \\
\hline
\end{tabular}
  \label{tab: MixVPR diferent M}
   \vspace{-0.2cm}
\end{table}

\begin{table}[t]
  \centering
\renewcommand\arraystretch{1.2}
\renewcommand\tabcolsep{8pt}
\captionsetup{aboveskip=-1pt, belowskip=-1pt}
  \caption{Study on different Data Organization Methods with MixVPR}
  \begin{tabular}{l|ccc}
  \hline
Data Organization Method &Cambridge    &LIB   &CON    \\
    \hline
Zhou~\etal~\cite{zhou2023nerf}  &68.39  &89.52 &92.14    \\
Zhou+Our-Data-Org. &\underline{71.06}  &\underline{91.97}  &\underline{93.24}  \\    
 \hline 
 Chen~\etal~\cite{chen2022dfnet}  &68.29 &89.36 &92.06     \\
Chen+Our-Data-Org. &\underline{71.58}  &\underline{92.14}  &\underline{93.32}  \\    
 \hline 
 Feldmann~\etal~\cite{feldmann2024nerfmentation}  &68.34  &88.86 &91.82     \\
Feldmann+Our-Data-Org. &\underline{71.32}  &\underline{93.04}  &\underline{93.24}  \\    
 \hline 
Ours &\textbf{72.05}  &\textbf{93.20}  &\textbf{93.71} \\
 \hline 
\end{tabular}
  \label{tab:data}
\vspace{-0.4cm}
\end{table}

\subsubsection{Impact of NeRF Performance}
In this study, we discuss the impact of different NeRF models on our method. In previous experiments, we used NeRF-H for image rendering. NeRF-H is an improved version of NeRF-O~\cite{mildenhall2021nerf}, capable of handling varying lighting conditions, occlusions, and image noise, while also providing more refined exposure control. 

As shown in~\cref{tab:Nerf on differnet dataset}, our method consistently outperforms the baseline in terms of Recall@1 across different NeRF models. The PSNR metric measures the rendering performance of NeRF methods. we present a comparison between the rendered images from different NeRF models and the corresponding real images in~\cref{fig:rendering}. We find an interesting correlation: higher PSNR values tend to be associated with better Recall@1 results. This is reasonable, as higher PSNR indicates better quality of synthetic images, which in turn enhances VPR performance.
\begin{table}[t]
\centering
\renewcommand\arraystretch{1.2}
\renewcommand\tabcolsep{1.5pt}
\captionsetup{aboveskip=--0.5pt, belowskip=-0.5pt}
\caption{Recognition performance with Different NeRF}
\begin{tabular}{l@{\hspace{0.2em}}|@{\hspace{0.2em}}c@{\hspace{0.2em}}c@{\hspace{0.2em}}c@{\hspace{0.2em}}c}\hline

Dateset &PSNR$\uparrow$ & MixVPR~\cite{ali2023mixvpr}    & EigenPlaces~\cite{Berton2023EigenPlaces}  & CricaVPR~\cite{lu2024cricavpr} \\
    \hline
Cam(No NeRF) &--  &67.50  &81.24  &53.08    \\
Cam(NeRF-O)  &19.83 &\textbf{72.10}
&\underline{89.29}
&\underline{59.46}   \\
Cam(NeRF-H)  &18.82
&\underline{72.05}
&\textbf{89.81} &\textbf{61.60}   \\
\hline 
LIB(No NeRF)  &-- &89.19  &92.30  &88.04    \\
LIB(NeRF-O)  &22.98
 &\underline{92.14}  &\underline{93.12}  &\underline{88.29}  \\
LIB(NeRF-H)  &25.71
 &\textbf{93.20} &\textbf{93.69}  &\textbf{88.53}  \\  \hline  
CON(No NeRF)  &-- &91.67&94.73 &89.54    \\
CON(NeRF-O)  &23.05
 &\underline{93.00}  &\textbf{95.99}
  &\underline{90.32}  \\
CON(NeRF-H)  &25.51
&\textbf{93.71} &\underline{95.28} &\textbf{90.64}  \\ 
\hline
\end{tabular}
  \label{tab:Nerf on differnet dataset}
  \vspace{-0.4cm}
\end{table}
\begin{figure*}[t]
  \centering
  \includegraphics[width=0.9\textwidth]{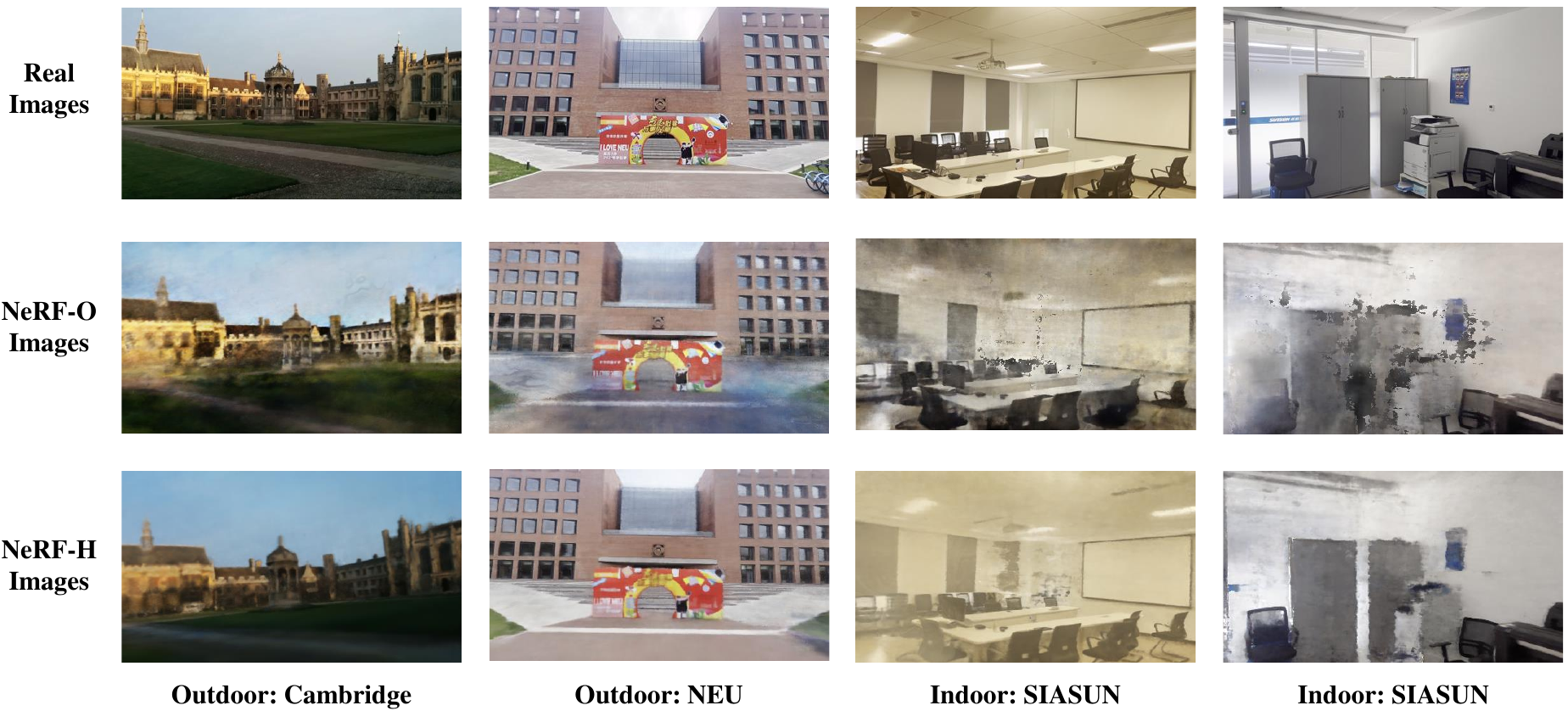}
  \captionsetup{aboveskip=2pt, belowskip=0pt}
  \caption{Rendering of the same place in different NeRF models.}
 \label{fig:rendering}
 \vspace{-0.7cm}
\end{figure*}
\section{Conclusion}
In this paper, we propose a training paradigm for VPR based on uncertainty-guided NeRF augmentation, UGNA-VPR, which improves VPR performance without modifying the VPR network structure or requiring additional data collection. We conducted extensive experiments on multiple indoor and outdoor datasets using various VPR network backbones and compared our method with other baseline training paradigms. The experimental results demonstrate that our method makes more effective use of existing available data, significantly enhancing VPR network training compared to baseline methods. Additionally, we propose a novel training data organization technique that scientifically manages both existing and NeRF-generated data, which is a good match with our UGNA-VPR and further improves VPR training effectiveness. We also conducted ablation studies to validate the effectiveness of all our designs. 

\bibliographystyle{IEEEtran}

\bibliography{manuscript}

\end{document}